\documentclass[conference]{IEEETran}
\IEEEoverridecommandlockouts
\usepackage{cite}
\usepackage{amsmath,amssymb,amsfonts}
\usepackage{algorithmic}
\usepackage{booktabs}
\usepackage{graphicx}
\usepackage{tabularx}
\usepackage{caption}
\usepackage{textcomp}
\usepackage{xcolor}
\def\BibTeX{{\rm B\kern-.05em{\sc i\kern-.025em b}\kern-.08em
    T\kern-.1667em\lower.7ex\hbox{E}\kern-.125emX}}
    
\graphicspath{{./figures/}}

\begin{document}

\title{Synaptic Stripping: How Pruning Can Bring Dead Neurons Back To Life}

\author{\IEEEauthorblockN{Tim Whitaker}
\IEEEauthorblockA{\textit{Department of Computer Science} \\
\textit{Colorado State University}\\
Fort Collins, CO, USA \\
timothy.whitaker@colostate.edu \\[-3.0ex]}
\and
\IEEEauthorblockN{Darrell Whitley}
\IEEEauthorblockA{\textit{Department of Computer Science} \\
\textit{Colorado State University}\\
Fort Collins, CO, USA \\
whitley@cs.colostate.edu \\[-3.0ex]}
}

\maketitle

\begin{abstract}
Rectified Linear Units (ReLU) are the default choice for activation functions in deep neural networks. While they demonstrate excellent empirical performance, ReLU activations can fall victim to the dead neuron problem. In these cases, the weights feeding into a neuron end up being pushed into a state where the neuron outputs zero for all inputs. Consequently, the gradient is also zero for all inputs, which means that the weights which feed into the neuron cannot update. The neuron is not able to recover from direct back propagation and model capacity is reduced as those parameters can no longer be further optimized. Inspired by a neurological process of the same name, we introduce Synaptic Stripping as a means to combat this dead neuron problem. By automatically removing problematic connections during training, we can regenerate dead neurons and significantly improve model capacity and parametric utilization. Synaptic Stripping is easy to implement and results in sparse networks that are more efficient than the dense networks they are derived from. We conduct several ablation studies to investigate these dynamics as a function of network width and depth and we conduct an exploration of Synaptic Stripping with Vision Transformers on a variety of benchmark datasets.
\end{abstract}

\section{Introduction}

Over the last decade, the rectified linear unit (ReLU) has become one of the most successful and widely used activation functions in neural networks. 
Defined as $f(x) = max(0, x)$, rectified linear units played a large role in the success of deep learning as they enabled neural networks to become deeper without falling prey to the vanishing gradient problem.
This happens with traditional sigmoidal activations, because they saturate when input values become larger.
Rectifiers instead have a positive unbounded range and constant gradient which avoids this problem and has been shown to significantly accelerate learning in deep neural networks \cite{NIPS2012_c399862d}.

Dead neurons are a common and well documented side effect that arises from the use of ReLU activation functions in deep neural networks \cite{Lu_2020, https://doi.org/10.48550/arxiv.1806.06068, https://doi.org/10.48550/arxiv.1812.05981}. 
As ReLUs always output zero for a negative number, it's possible that the weights for a neuron can be distributed such that no reasonable input can activate that neuron.
If a neuron falls into this state, the gradient becomes zero for all training inputs and the neuron's weights cannot be directly updated by back propagation.
Dead neurons reduce model capacity and parametric utilization since the weights feeding into a dead neuron are unable to be optimized during training.


Several rectifier variations have been introduced to address the dead neuron problem, including:
Leaky ReLU, Parametric ReLU, Exponential Linear Units, and Gaussian Error Linear Units  \cite{https://doi.org/10.48550/arxiv.1502.01852, https://doi.org/10.48550/arxiv.1511.07289,https://doi.org/10.48550/arxiv.1606.08415}.
These functions contain non-linear or non-constant outputs in parts of their negative domain which results in non-zero gradients that ReLU would otherwise have.
However, dataset and network architecture can significantly influence the performance of activation functions, and it's not conclusive that these variations outperform ReLU in many cases \cite{https://doi.org/10.48550/arxiv.2109.14545}.
Alternative initialization schemes and training methodologies have also been introduced to try to alleviate the dead neuron problem, which include lower learning rates, reparameterization of dead neurons, and asymmetric weight distributions for initialization.
These methods can reduce the number of dead neurons, but they can also negatively impact the speed and quality of convergence.

Inspired by a biological process of the same name, we introduce \textit{Synaptic Stripping} as an effective method for regenerating dead neurons by selectively pruning problematic connections.
Paradoxically, we demonstrate that removing parameters from a network can actually increase model capacity without adversely affecting training or generalization performance.

We explore the efficacy of Synaptic Stripping with Vision Transformers on a variety of benchmark datasets.
Vision Transformers are quickly growing in popularity as they continue to produce state-of-the-art results on difficult image classification tasks \cite{zhai2021scaling}.
We conduct a large scale comparison of Synaptic Stripping using both GELU and ReLU activation functions on Tiny Imagenet, SVHN, CIFAR-10, and CIFAR-100.
We also conduct several ablation studies using multilayer perceptrons in order to probe the dynamics of dead neurons in networks of varying widths and depths.

Our results indicate that Synaptic Stripping can significantly increase the model capacity of large scale vision transformers by up to 30\% while improving generalization and robustness on in-distribution and out-of-distribution datasets.
Synaptic Stripping results is significant improvements over baseline models that use ReLU while producing slightly improved results over GELU.
Our method results in better utilization of network parameters, decreased model size, and improved training efficiency with minimal computational cost.

\begin{figure*}
    \centering
    \includegraphics[width=\textwidth]{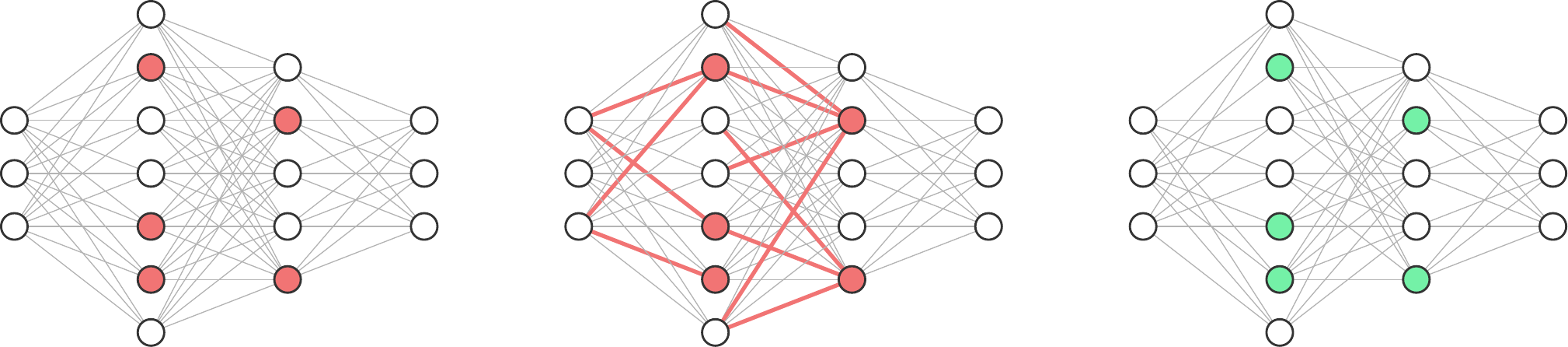}
    \caption{An illustration of Synaptic Stripping. After each training epoch, dead neurons are detected. Problematic connections associated with dead neurons are pruned. The same dead neurons now become active and training continues.}
\end{figure*}

\pagebreak

\section{Background}

\subsection{Neuroscience Observations}

Sparse network structures and neuroregeneration play a critical role in healthy brain function.
One early process in brain development involves a massive proliferation and subsequent apoptosis, which results in the pruning of billions of neurons and trillions of synaptic connections.
The neuronal proliferation creates an environment in which neurons and connections compete for resources.
Invariably, many neurons and connections die and the resulting network becomes highly specialized, consisting of many sparsely connected neuronal groups.
These neuronal groups then form the primary repertoire from which the rest of brain development occurs \cite{edelman1987neural}.

Under metabolic energy constraints, this overgrowth and subsequent pruning has been shown to maximize memory performance \cite{10.1162/089976698300017124}.
Additionally, this neuronal connectivity allows for information to be encoded in a sparse and distributed manner, which enables networks to optimize a tradeoff between energy expenditure and expressivity \cite{doi:10.1097/00004647-200110000-00001, pmlr-v15-glorot11a}.


Neuroregeneration is a growing field that encompasses repair and regrowth of the nervous system.
One important process called \textit{Synaptic Stripping} has been documented as playing an important role in neuroregeneration, where immunocompetent cells called microglia constantly scan the brain and selectively remove dysfunctional synapses from injured neurons \cite{KETTENMANN201310}.
There is significant potential impact in better understanding how dead neurons and neuroregeneration affect neural systems.

\subsection{The Dying ReLU Problem}

\begin{figure*}
    \centering
    \includegraphics[width=\textwidth]{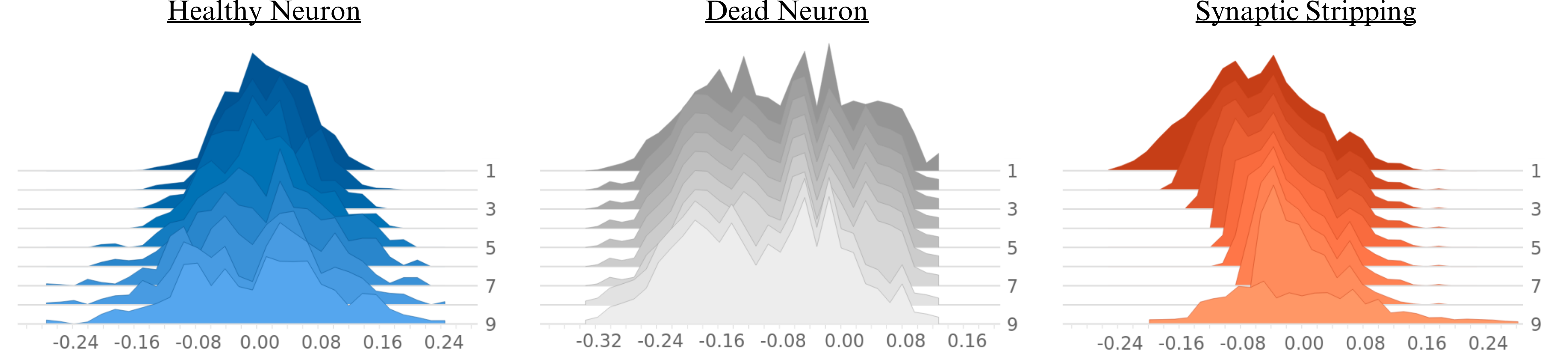}
    \caption{A visualization of neuron weights evolving over time. We display histogram snapshots of the weights of three neurons after each epoch of training. To illustrate how Synaptic Stripping works, we include a healthy neuron, a dead neuron, and a dead neuron that is trained with Synaptic Stripping. After several epochs of pruning negative weights, Synaptic Stripping successfully regenerates the dead neuron, enabling further optimization.}
    \label{fig:my_label}
\end{figure*}

Rectified Linear Units (ReLU) are non-linear activation functions that return the input if it is positive and zero otherwise.
ReLU is cheap to compute in both the forward and backward pass, and the sparse activation patterns have been theorized to allow for better information disentangling \cite{pmlr-v15-glorot11a}.

However, ReLU units can suffer from the dead neuron problem, where the weights for a given neuron are pushed into a state where the output of the ReLU is zero for any reasonable input.
This can be illustrated by considering a standard, fully connected layer that contains some number of neurons with ReLU activation functions. Each neuron computes a weighted sum of all of the inputs $x$ with weights $w$ and a bias term, which is then passed through the activation function.
\[
f(x) = ReLU(\sum_{i=1}^m w_i x_i + bias)
\]

\[
ReLU(x) = \begin{cases}
x & \text{if} \ \ x > 0 \\
0 & \text{otherwise}
\end{cases}
\]

To illustrate a worst case scenario, assume that the bias term is insignificant.
If all inputs $x$ into a given neuron are positive and all weights are negative, then for any given input the output of the ReLU activation will be zero.
Consequently, when we back propagate, no updates will occur to the weights of a dead neuron since the derivative of the ReLU function is zero for inputs less than zero.

The dying ReLU problem is well documented, and many researchers have introduced methods for alleviating it in deep neural networks. 
Perhaps the most prevalent reason for neurons being pushed into these dead states is large weight updates as a result of large learning rates or momentum \cite{phdthesis}.
The simple solution is to use small learning rates, however this can drastically reduce training efficiency.
In practice, many models today implement learning rate schedules that use large learning rates at the beginning of training which then decay over time \cite{you2019lrdecay}.
In these cases, dead neurons develop early and persist through the rest of training.
Additionally, adaptive optimizers are commonly used, which use running averages to update momentum and learning rates for each parameter individually throughout training \cite{adam}.
These adaptive optimizers significantly speed up training, but they cause large jumps that can push neurons into dead states.

Some work has investigated how initial weight distributions impact the probabilities of ReLU neurons being born dead \cite{Lu2019DyingRA}.
Typically, weights are initialized using random symmetric distributions centered around a zero mean.
As networks get deeper and narrower, these symmetric distributions can lead to network collapse as the asymptotic probability of dead neurons causing a bottleneck approaches one.
However, since ReLU is an asymmetric function, the probability bounds for born dead neurons can be drastically reduced by using a random asymmetric initialization \cite{Lu2019DyingRA}.

It's also been shown that reinitializing dead neurons can have significant impact on performance.
Neural Rejuvenation was introduced as an optimization method for convolutional networks that reparameterizes and reallocates dead neurons during training \cite{https://doi.org/10.48550/arxiv.1812.00481}.
Here they define dead neurons to be convolutional channels that have a low impact on the output.
Low impact channels are removed from the given layer, moved to another layer and reinitialized.
In some respects, this method is a neural architecture search algorithm that finds layer size configurations while optimizing parameters at the same time.

\subsection{Activation Function Trends in Computer Vision}

Logistic Sigmoids and Hyperbolic Tangents were the most widely used activation functions in early computer vision research. In 2012, the seminal AlexNet paper led to the popularity and wide adoption of ReLU as activation functions in deep neural networks \cite{NIPS2012_c399862d}.
In 2015, residual networks utilized skip connections and ReLU activations to enable extremely deep networks \cite{resnet}.
ResNets have been extremely influential and have been the building block for many computer vision architectures over the last several years.

Several new activation functions were introduced to improve on ReLU.
Leaky ReLUs, Parametric ReLUs, Exponential Linear Units, Gaussian Error Linear Units, and Sigmoid linear Units contain non-linear or non-constant activations in the negative parts of their domain, so that neurons could potentially recover if they end up in a state that would be considered dead with ReLUs \cite{https://doi.org/10.48550/arxiv.1502.01852, https://doi.org/10.48550/arxiv.1511.07289,https://doi.org/10.48550/arxiv.1606.08415, https://doi.org/10.48550/arxiv.1710.05941, https://doi.org/10.48550/arxiv.1908.08681}.
However, despite the purported benefits of these alternatives, there is not conclusive evidence that they outperform ReLU.
Dataset and network architecture can influence performance and research has shown that ReLU is competitive and even outperforms these other activations on benchmark datasets with convolutional networks \cite{https://doi.org/10.48550/arxiv.2109.14545}.


\subsection{Transformers}

In natural language processing, transformers now dominate the field \cite{NIPS2017_3f5ee243}.
These models make use of attention mechanisms and linear layers with Gaussian Error Linear Units (GELU).
Recently, these architectures have made their way to computer vision and have produced several state of the art results on large datasets \cite{DBLP:conf/iclr/DosovitskiyB0WZ21}. 
Borrowing from their natural language variants, vision transformers also use GELU functions. 
However, there has been limited work exploring the efficacy of different activation functions with these models.
Activation function performance can depend on dataset and architecture choice \cite{https://doi.org/10.48550/arxiv.2109.14545}.
So while there are claims of GELU being more effective on language tasks with transformers, there is not conclusive evidence that this assumption holds on image datasets with vision transformers.

\section{Synaptic Stripping}

Our approach regenerates dead ReLU neurons by selectively pruning synaptic connections.
Synaptic Stripping can be plugged into any standard training loop.
We begin by initializing and training a neural network using a standard optimization algorithm and learning rate schedule.
After each training epoch, we evaluate the network on a validation set, while keeping track of the outputs of each ReLU.
Each neuron that has a total activation of zero over the entire validation set is considered dead.
As ReLU is a positive asymmetric function, we prune the most negative weights feeding into dead neurons, shifting weight distributions towards a more positive mean.
Training then continues with regenerated neurons, and this process is repeated until convergence.

\subsection{Dense Layers vs Convolutional Layers}

We focus our research on dead neurons which are located in dense, fully connected layers. While convolutional layers do suffer from the dead neuron problem, the local connectivity and weight sharing aspect of convolutional filters makes pruning a less appealing option for regenerating dead units.

In convolutional layers, filters slide over the input to produce feature maps.
Pruning weights from a filter impacts not only the neuron that corresponds to that location in the feature map, but also all other neurons in the channel.
With dense layers, neurons are fully connected and independent of all other neurons in the same layer.
Therefore, pruning can target dead neurons without adversely affecting other neurons in the same layer.
Additionally, pruning dead neurons leads to a strict increase in capacity.
Since the weights feeding into a dead neuron are not being utilized, there is no danger of reducing capacity by removing them.
In the worst case, pruning a small number of dead neuron weights results in a neuron that continues to stay dead and functional capacity is identical. 
In the best case, some small percentage of connections are removed, the neuron is regenerated, and the model can now make use of the extra parameters feeding into the neuron that were not pruned.

\subsection{Detecting Dead Neurons}

Dead neurons can be identified by analyzing the outputs of neurons in the forward pass or the gradients of neurons in the backward pass.

Gradient based detection is closely related to saliency based pruning methods \cite{NIPS1989_6c9882bb, https://doi.org/10.48550/arxiv.2006.05467}.
The goal is to identify neurons or parameters that have the lowest impact on the network's output.
This distinction can lead to confusion between neurons that are dead (always output zero) and neurons that are frozen (not being updated) \cite{phdthesis}.
As gradients are calculated by differentiating the error with respect to the weights, a low error can result in a zero gradient for a specific neuron even though it is active.
This can happen when the network exists in some local optima or the weights for a neuron exist in some flat region of the loss landscape.
The ambiguity between frozen neurons and dead neurons makes gradient based detection a poor fit for our method.

We instead detect dead neurons by analyzing the output activations of neurons on the forward pass.
This approach is simple in practice since we can just keep a running sum of each neuron's outputs over all batches or samples in the dataset.
Any neuron that has a sum total activation of zero is considered dead.
This constraint can be relaxed with a small threshold to allow for neurons with very low activations in some cases.
In these cases, a neuron may be considered "mostly dead" if it is only active for a couple of samples in a large dataset.
Synaptic Stripping may be beneficial in encouraging mostly dead neurons to become more active.

Dead neurons can be evaluated over an entire training set or over a validation set.
Since validation sets are much smaller than training sets, it's possible that dead neurons found over a validation set would not appear as dead neurons during training.
In these cases, we would be pruning neurons that have potential for further optimization.
However, as these neurons are dead for all predictions that the model is evaluated on, then validation set detection could offer some form of regularization over a more robust training set detection. Unless otherwise stated, our experiments all use validation set detection with a standard 80/10/10 train/validation/test split.

\subsection{Pruning Methodology}

We implement Synaptic Stripping by removing some of the most negative weights from each dead neuron at the end of each training epoch.
The amount of weights to remove depends on the network architecture, the training data, the number of training epochs, and the frequency of pruning.

In our experiments, we prune ten percent of the weights feeding into each dead neuron at each iteration.
It is possible that ten percent is not enough to regenerate certain neurons in one shot. 
Thus, the same neuron may need to be pruned over multiple epochs in order to regenerate.
We find that this approach prevents over-pruning, which would reduce total capacity by pruning more weights than is necessary.

The effects of Synaptic Stripping can be visualized by looking at histograms of the weight distributions of dead neurons over time.
Figure 2 shows how weight distributions evolve by recording neuron weights taken at the end of each epoch for ten epochs.
Healthy neurons end up more or less centered around a zero mean as weights drift towards their optima.
The dead neuron's weights are strongly skewed towards negative values at the first epoch and over time they remain frozen.
This negative skew is typical of dead neurons in networks that we analyzed.
When Synaptic Stripping is implemented, a small percentage of the most negative weights feeding into the dead neuron are removed at each epoch, which shifts the distribution in the positive direction so that the neuron is able to regenerate.

\subsection{Model Capacity}

Complexity is a fundamental problem in machine learning theory, where the goal is to understand the expressivity and learnability of different models.
Some popular complexity metrics introduced include Vapnik–Chervonenkis (VC) dimension and Rademacher complexity \cite{doi:10.1137/1116025, bartlett2002rademacher} which provide loose bounds on a model's ability to shatter a dataset and to fit random labels respectively.
However, since these metrics provide only loose bounds and modern deep neural networks are extremely overparameterized, this results in values that are a poor fit for meaningful comparison \cite{hu2021complexity}.

The simplest and most common way to consider model capacity is to simply count the number of parameters.
In our case, it is more interesting to consider the number of \textit{active} parameters a model contains.
Since dead neurons always output zero and are incapable of further training or optimization, they in turn have no impact on subsequent computation.
Dead neurons and all associated input and output weights can then be removed from a network with no effect on functional behavior.
So in order to compute the number of active parameters, we simply count the number of parameters that are not attached to a dead neuron.
For networks trained with Synaptic Stripping, we count the number of parameters that are not attached to dead neurons, while adjusting for the number of pruned connections.

This provides a simple and clean way to compare the capacity between a network that has dead neurons and a network that has been pruned.

\section{Experiments}

\begin{figure}
    \centering
    \includegraphics[width=\columnwidth]{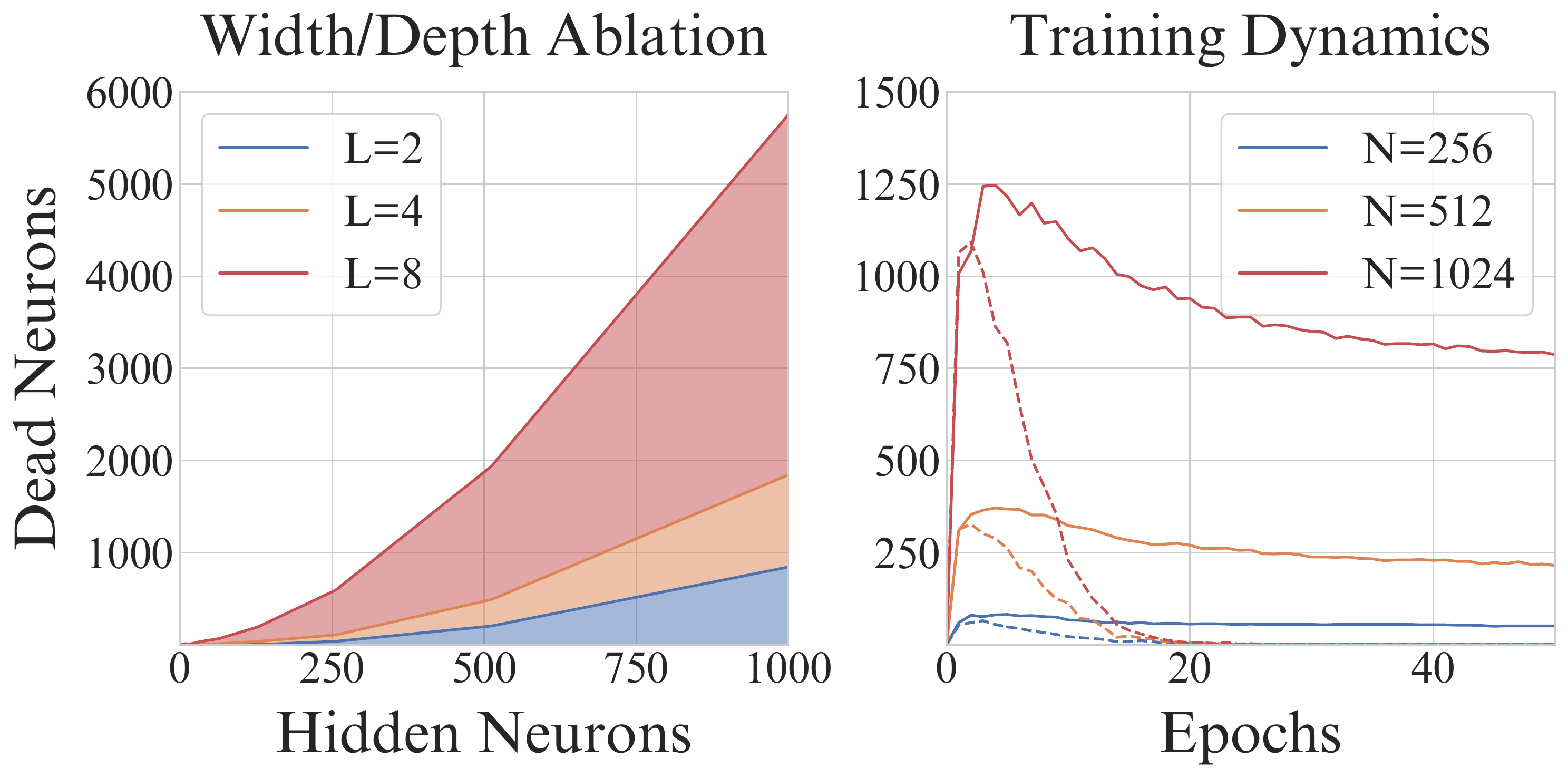}
    \caption{Explorations of dead neurons with multilayer perceptrons on CIFAR-10.
    The leftmost graph shows the number of dead neurons (w/o Synaptic Stripping) increasing as a function of both model width and depth.
    The rightmost graph shows how dead neurons develop over 50 epochs of training with a 2 layer network of varying widths.
    The results for Synaptic Stripping are shown as dashed lines, and the corresponding results without pruning as solid lines. }
    \label{fig:my_label}
\end{figure}

\textbf{Datasets}: We use the computer vision datasets, Tiny Imagenet \cite{Le2015TinyIV}, CIFAR-10 CIFAR-100 \cite{Krizhevsky09learningmultiple}, and SVHN \cite{37648}.

Tiny ImageNet is a subset of the famous ImageNet dataset with images scaled and downsized to 64x64 pixels. Tiny ImageNet contains 100,000 images of 200 classes, where each class contains 500 training images, 50 validation images, and 50 test images. We perform an additional downsize to 32x32 pixels to keep our models consistent between all experiments on CIFAR and SVHN.

CIFAR consists of 60,000 small natural colored images of 32x32 pixels in size. Those 60,000 images are split up into 50,000 training images and 10,000 testing images. CIFAR-10 samples from 10 classes of images, while CIFAR-100 samples from 100 classes of images. CIFAR-100 is more difficult than CIFAR-10 as each class will have only 500 training samples compared to 5,000 in CIFAR-10.

SVHN is the Street View House Number dataset which contains 600,000 32x32 real world images of house numbers obtained from Google's Street View cameras. The goal is to classify single digits (0 to 9). The images are cropped so that a single digit is centered, but many photos contain other digits and distractors as well. The dataset is split into a training set of 73,257 images and a testing set of 26,032 images. The remaining 531,131 images are included as additional examples to use as extra training data. We do not use these extra images in our experiments.

We also test our methods on corrupted variations of all four of the previous datasets. These datasets are created by applying up to 20 different distortions (gaussian noise, blur, pixelation, snow, etc.) at five different severity levels \cite{https://doi.org/10.48550/arxiv.1903.12261}.
These additional test sets allow us to measure the robustness of models on out-of-distribution data, which allows us to get a better sense of generalization performance.

\textbf{Models}: We use a standard multilayer perceptron for our ablation experiments. These models consist only of fully connected layers followed by ReLU activations. For our large scale experiment, we use an open source implementation of the vision transformer with hyperparamters selected to be appropriate for small scale datasets \cite{Omiita}. All models are implemented and tested with PyTorch.

\textbf{Optimization}: We use Adaptive Moment Estimation (Adam) for all of our experiments. This optimizer keeps track of an exponentially decaying moving average of previous gradients and squared gradients so that individual parameters are updated with adapted learning rates. Adam is known for its training efficiency compared to stochastic gradient descent and is commonly used across many machine learning tasks \cite{DBLP:journals/corr/KingmaB14}.

\textbf{Data Augmentation}: Transformers need larger amounts of data to perform well without overfitting. Data augmentation is commonly used as a way to artificially expand the size of training sets. We use AutoAugment for our vision transformer experiments, which is a collection of optimal augmentation policies for CIFAR, SVHN, and ImageNet found by searching through a space of various image perturbations including operations like translation, rotation, shear, contrast, etc \cite{https://doi.org/10.48550/arxiv.1805.09501}.

\textbf{Hardware}: All models are trained on a single Nvidia GTX-1080-Ti GPU.

\begin{table}
\small
\caption*{CIFAR-10}
\begin{tabularx}{\columnwidth}{c c c c c c@{\hspace{0.05cm}} c c c}
\toprule
& & \multicolumn{3}{c}{Baseline} & & \multicolumn{3}{c}{Synaptic Stripping} \\
\cmidrule(lr){3-5} \cmidrule(r){7-9}
L & N & Acc $\uparrow$ & \#D $\downarrow$ & \%A $\uparrow$ & & Acc $\uparrow$ & \#D $\downarrow$ & \%A $\uparrow$ \\
\midrule
2 & 256 & 53.28 & 30 & 88.62 & & 53.74 & 0 & 94.01 \\
2 & 512  & 53.89 & 224 & 61.04 & & 53.94 & 0 & 82.88  \\
2 & 1024 & 53.92 & 1010 & 25.69 & & 54.38 & 3 & 68.67  \\
\midrule
4 & 256 & 53.50 & 63 & 88.07 & & 53.54 & 1 & 91.23  \\
4 & 512 & 54.27 & 645 & 46.93 & & 54.42 & 5 & 83.02  \\
4 & 1024 & 55.04 & 2610 & 13.16 & & 54.67 & 53 & 59.07  \\
\midrule
8 & 256 & 52.72 & 781 & 38.58 & & 52.92 & 6 & 73.95 \\
8 & 512 & 53.89 & 2566 & 13.95 & & 53.89 & 31 & 64.33  \\
8 & 1024 & 54.08 & 6052 & 6.82 & & 54.04 & 248 & 43.57  \\
\bottomrule
\end{tabularx}

\caption*{CIFAR-100} 
\vspace{0.1in}
\begin{tabularx}{\columnwidth}{c X c c c c@{\hspace{0.05cm}} c c c}
\toprule
& & \multicolumn{3}{c}{Baseline} & & \multicolumn{3}{c}{Synaptic Stripping} \\
\cmidrule(lr){3-5} \cmidrule(r){7-9}
L & N & Acc $\uparrow$ & \#D $\downarrow$ & \%A $\uparrow$ & & Acc $\uparrow$ & \#D $\downarrow$ & \%A $\uparrow$\\
\midrule
2 & 256 & 23.33 & 51 & 81.07 & & 23.53 & 1 & 92.39  \\
2 & 512  & 23.01 & 215 & 62.26 & & 23.24 & 0 & 81.48  \\
2 & 1024 & 22.57 & 787 & 37.91 & & 22.43 & 0 & 72.48  \\
\midrule
4 & 256 & 23.03 & 97 & 81.95 & & 23.34 & 2 & 90.06 \\
4 & 512 & 23.06 & 539 & 54.29 & & 23.09 & 2 & 81.15  \\
4 & 1024 & 22.40 & 1970 & 26.94 & & 22.82 & 1 & 62.22 \\
\midrule
8 & 256 & 20.68 & 557 & 53.00 & & 21.00 & 5 & 75.87 \\
8 & 512 & 21.14 & 1975 & 26.81 & & 21.37 & 6 & 67.18 \\
8 & 1024 & 20.58 & 5454 & 11.17 & & 21.84 & 78 & 47.98 \\
\bottomrule
\end{tabularx}
\caption{Results from the ablation experiments with MLPs of varying hidden depths and widths. All models trained for 50 epochs with Adam. Metrics include (L) number of hidden layers, (N) number of hidden neurons, (Acc.) peak validation accuracy, (\#D) number of dead neurons, and (\%A) percentage of active parameters.}
\end{table}

\subsection{MLP Ablations}

\begin{figure*}
    \centering
    \includegraphics[width=\textwidth]{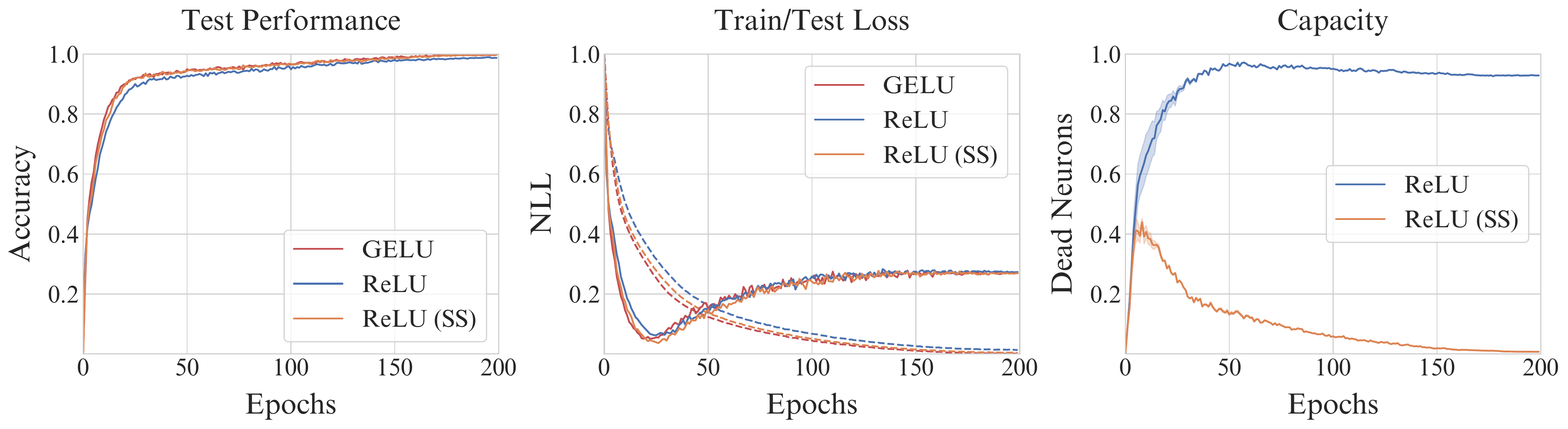}
    \caption{Mean normalized accuracy, loss, and capacity for vision transformers over all four datasets (CIFAR-10, CIFAR-100, SVHN, Tiny Imagnet). Training loss is indicated with a dashed line. Test loss with a solid line. Performance is very similar between ReLU with Synaptic Stripping (``ReLU (SS)") and GELU.  There is a significant increase in performance and capacity with Synaptic Stripping over the baseline ReLU.}
    \label{fig:my_label}
\end{figure*}

\begin{table*}[h]
\small
\centering
\parbox{120pt}{
\begin{tabularx}{120pt}{c c}
\toprule
Parameter & Value \\
\midrule
\addlinespace[0.4em]
Epochs & 200 \\
\addlinespace[0.02em]
Warmup Epochs & 5 \\
\addlinespace[0.02em]
Batch Size & 128 \\
\addlinespace[0.02em]
Optimizer & Adam \\
\addlinespace[0.02em]
Scheduler & Cosine \\
\addlinespace[0.02em]
$\alpha_1$, $\alpha_2$ & 1e-3, 1e-5 \\
\addlinespace[0.02em]
$\beta_1$, $\beta_2$ & 0.9, 0.999 \\
\addlinespace[0.02em]
Weight Decay & 5e-5 \\
\addlinespace[0.02em]
Patch Size & 8 \\
\addlinespace[0.02em]
Heads & 8 \\
\addlinespace[0.02em]
Layers & 7 \\
\addlinespace[0.02em]
Hidden Size & 384 \\
\addlinespace[0.02em]
Expansion & 4 \\
\addlinespace[0.1em]
\bottomrule
\end{tabularx}
}
\hspace{20pt}
\parbox{325pt}{
\begin{tabularx}{325pt}{c c c c c c c c}
\toprule
Activation & Dataset & Acc $\uparrow$  & NLL  $\downarrow$ &  cAcc $\uparrow$ & cNLL $\downarrow$ & \#D  $\downarrow$ & \%A  $\uparrow$ \\
\midrule
GeLU & CIFAR-10 & 89.27 & 0.423 & 79.86 & 0.912 &  &  \\
ReLU &  & 88.75 & 0.425 & 79.14 & 0.946 & 1511 & 64.86 \\
ReLU (SS) & & 89.29 & 0.419 & 80.07 & 0.898 & 9 & 87.12 \\
\midrule
GeLU & CIFAR-100 & 61.62 & 1.724 & 50.36 & 3.141 & & \\
ReLU &  & 61.27 & 1.729 & 49.39 & 3.170 & 1051 & 75.56 \\
ReLU (SS) & & 61.57 & 1.733 & 50.17 & 3.151 & 2 & 89.58 \\
\midrule
GeLU & SVHN & 97.35 & 0.118 & 92.44 & 0.377 & & \\
ReLU &  & 97.08 & 0.131 & 92.26 & 0.446 & 2001 & 53.47 \\
ReLU (SS) &  & 97.37 & 0.118 & 93.13 & 0.376 & 4 & 87.26 \\
\midrule
GeLU & Tiny Imagenet & 38.75 & 2.738 & 26.67 & 6.617 &  &  \\
ReLU &  & 38.81 & 2.735 & 26.21 & 6.577 & 995 & 76.86 \\
ReLU (SS) &  & 38.87 & 2.736 & 26.61 & 6.656 & 2 & 90.25 \\
\bottomrule
\end{tabularx}
}
\caption{Hyperparameters and results for vision transformers comparing activation functions and Synaptic Stripping. "ReLU (SS)" include Synaptic Stripping.
All models are trained twice and the best runs are reported. Metrics include (Acc) accuracy, (NLL) negative log likelihood, (\#D) number of dead neurons, and (\%A) percentage of active parameters. We also report (cAcc) accuracy and (cNLL) negative log likelihood for each final model evaluated on a larger corrupted dataset.}
\end{table*}

We begin by exploring how Synaptic Stripping affects simple models of varying width and depth on the CIFAR-10 and CIFAR-100 dataset. The model variations differ in the number of hidden layers between [2, 4, 8] and the number of neurons in each hidden layer between [256, 512, 1024].

Each model is trained for 50 epochs using the Adam optimizer with a learning rate of 1e-3 and a batch size of 64. All data is transformed with mean standard normalization with $\mu = $
(0.4914, 0.4822, 0.4465); $\sigma =$ (0.2023, 0.1994, 0.2010) and $\mu =$ (0.5071, 0.4867, 0.4408) $\sigma =$ (0.2675, 0.2565, 0.2761) for CIFAR-10 and CIFAR-100 respectively.

Each model is trained with and without Synaptic Stripping.
Synaptic Stripping is applied at each epoch for all dead neurons found through validation set detection.
We prune 10\% of the most negative weights for each dead neuron.
For neurons that have previously been pruned, Synaptic Stripping removes 10\% of the remaining weights.

For the baseline models and models trained with Synaptic Stripping, we keep track of the number of dead neurons and the number of active parameters.
The number of active parameters are calculated by subtracting the number of parameters connected to all dead neurons and the number of parameters pruned from the total number of hidden parameters in order to get a percentage value for the total capacity of the hidden layers.

Table 1 contains results for the ablation experiments.
For these simple models, accuracy does improve on both CIFAR-10 and CIFAR-100 for most model configurations, however the difference is small.
More importantly, Synaptic Stripping significantly improves model capacity across the board for all layer depth and width configurations.

\subsection{Vision Transformers}

\begin{figure}
    \centering
    
    \includegraphics[width=\columnwidth]{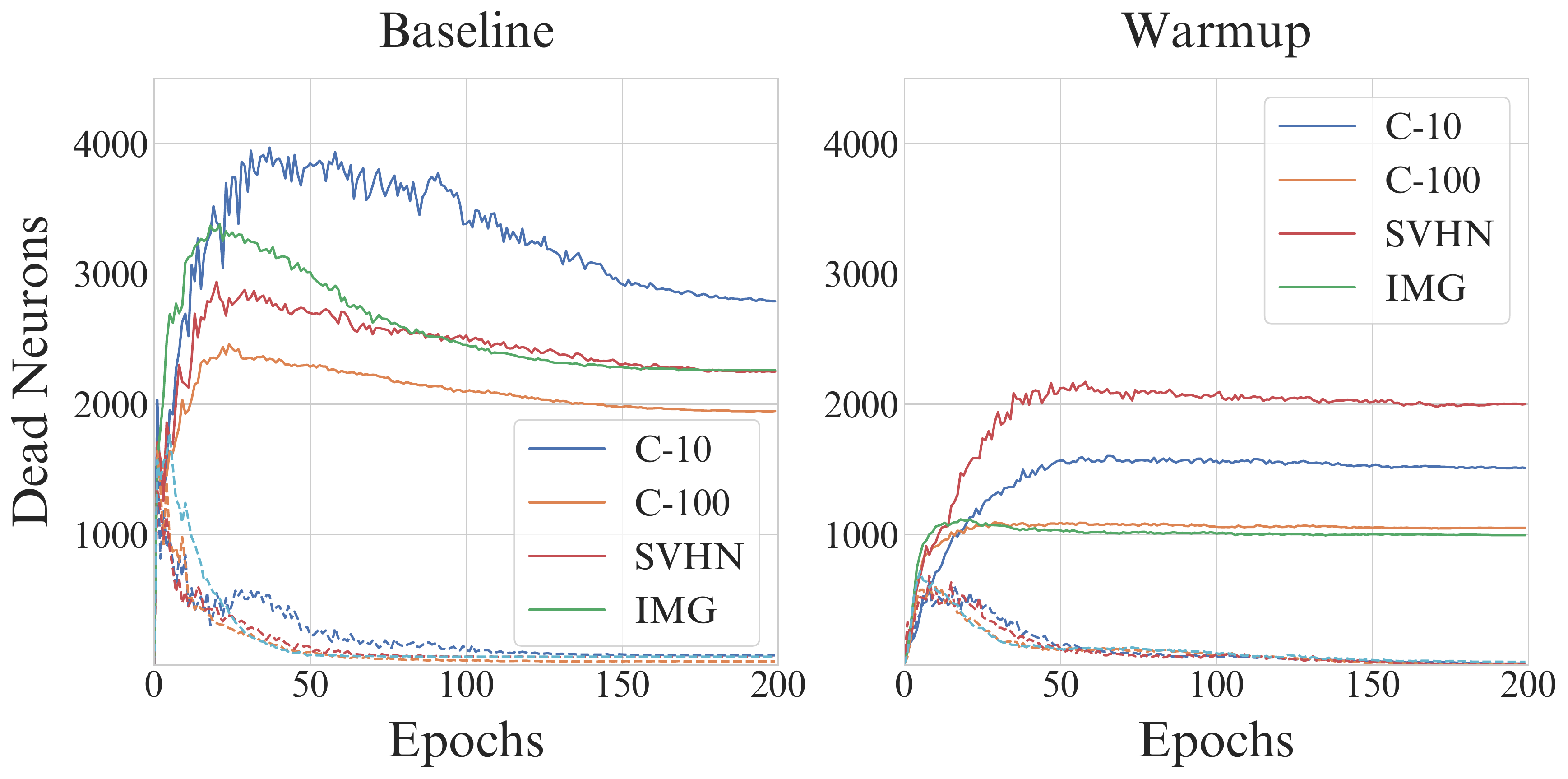}

    \vspace{0.2in}

    \small
    
    \begin{tabularx}{\columnwidth}{X c c c c c}
        \toprule
        & & \multicolumn{2}{c}{ReLU} & \multicolumn{2}{c}{ReLU (SS)} \\
        \cmidrule(r){3-4} \cmidrule(l){5-6}
        Schedule & Dataset & \#D $\downarrow$ & \%A $\uparrow$ & \#D $\downarrow$ & \%A $\uparrow$ \\
        \midrule
        Baseline & CIFAR-10 & 2787 & 35.20 & 72 & 79.11  \\
        Warmup & & 1511 & 64.86 & 9 & 87.12 \\
        \midrule
        Baseline & CIFAR-100 & 2138 & 50.28 & 26 & 81.42  \\
        Warmup & & 1051 & 75.56 & 2 & 89.58 \\
        \midrule
        Baseline & SVHN & 2249 & 47.71 & 58 & 83.63  \\
        Warmup & & 2001 & 53.47 & 4 & 87.26 \\
        \midrule
        Baseline & Tiny Imagenet & 2259 & 47.47 & 62 & 74.93 \\
        Warmup & & 995 & 76.86 & 2 & 90.25 \\
        \bottomrule
    \end{tabularx}

    \vspace{0.1in}
    
    \caption{Results detailing how warmup epochs affect dead neuron dynamics in Vision Transformers with and without Synaptic Stripping on CIFAR-10, CIFAR-100, SVHN, and Tiny Imagenet. The leftmost graph displays the number of dead neurons when the model is trained with a cosine annealing learning rate schedule from [1e-3, 1e-5]. The rightmost graph includes 5 epochs of linear learning rate warmup from [0, 1e-3] before then decaying from [1e-3, 1e-5]. Dashed lines represent models trained with Synaptic Stripping. Number of dead neurons (\#D) and percentage of active parameters (\%A) are reported for each case.}
\end{figure}

In addition to our baseline experiments, we explore Synaptic Stripping with vision transformers on several benchmark datasets. The original vision transformer was introduced for large datasets with several suggested configurations; the smallest (Vit-B) used 86 million parameters. We use an open source and standardized implementation with hyperparamters suitable for much smaller image sizes and datasets \cite{Omiita}.

We evaluate three models: one standard vision transformer with GELU activations, one with ReLU activations, and one with ReLU activations trained with Synaptic Stripping. All models contain seven transformer encoder layers and eight attention heads. Each transformer encoder contains a self attention block with a hidden size of 384 followed by a two layer MLP with an expansion factor of 4 for a hidden size of 1536. These MLP layers contain the activation functions and all Synaptic Stripping occurs with only these layers. We do not prune any of the attention modules. 

Each model is trained for 200 epochs using Adam with a batch size of 128. We incorporate a warmup phase for 5 epochs where the learning rate is linearly increased from 0 to 1e-3. This is known to reduce the large or divergent variance between runs \cite{https://doi.org/10.48550/arxiv.1910.04209}. We then use a cosine annealing learning rate schedule which decays from an initial rate of 1e-3 to a final rate of 1e-5 for the remaining 195 epochs. We train all models on CIFAR-10, CIFAR-100, SVHN and Tiny ImageNet. We test each model on the validation sets and the corrupted validation sets.

All images are sized to 32x32x3 and all datasets are expanded with AutoAugment policies. For Tiny ImageNet, we achieved better results using the CIFAR policy as opposed to the ImageNet policy due to the difference in image size that we use (32x32x3), as opposed to the original Imagenet dimensions (224x224x3).

    


We report peak validation accuracy, negative log likelihood, corrupted accuracy, corrupted negative log likelihood, number of dead neurons and percentage of active parameters. The number of dead neurons is not applicable for GELU as it does not suffer from the dead ReLU phenomena. We leave those fields blank in Table 2.

Our experiments indicate that ReLU with Synaptic Stripping significantly outperforms standard ReLU. Synaptic Stripping also outperforms GELU on all datasets except CIFAR-100; however, the performance gap between the two is extremely small. 
These patterns are consistent on the out-of-distibution corrupted datasets as well.
Synaptic Stripping is consistent at significantly reducing the number of dead neurons. 
On these small scale datasets, Synaptic Stripping increases capacity by 15\% to 33\% while reducing total parameter counts by 10\% to 15\% at the same time.
It's interesting to note that more dead neurons develop on datasets with fewer classes and more samples per class (CIFAR-10 and SVHN) compared to datasets with more classes and fewer samples per class (CIFAR-100 and Tiny Imagenet).

\textit{The Impact of Warming Up}: 
Our initial experiments used a standard cosine annealing training schedule without warmup epochs where we found a significant amount of variance between runs.
Since Adam calculates gradient statistics using a moving average, early epochs have a large impact on training, sometimes resulting in convergence to local optima with much worse performance.
After implementing a 5 epoch learning rate warmup schedule, our runs became much more consistent and with better accuracy.
Figure 5 details the results of our comparisons between runs with and without warmup epochs.
We find that runs with warmup epochs results in a significant reduction of the number of dead neurons and an increase in the percentage of active parameters in both the baseline models trained without Synaptic stripping as well as those trained with Synaptic Stripping.

\section{Conclusions}

We introduced Synaptic Stripping as an elegant way to regenerate dead rectifier neurons (ReLUs) in neural networks. 
Inspired by a neurological process of the same name, Synaptic Stripping is an iterative method that detects dead neurons, identifies problematic connections, and prunes those connections.  
After each training epoch, we detect dead neurons by calculating the sum of ReLU outputs for each neuron through a validation set. 
Since dead neurons are caused by a weight distribution that results in zero outputs, we prune some of the most negative weights from each of the dead neurons.
The strength and frequency of Synaptic Stripping can be tuned for a given model or dataset.
Our approach is flexible, extremely simple to implement for standard training methods and highly effective at regenerating dead neurons. 

We conduct a number of experiments exploring Synaptic Stripping with simple multilayer perceptrons and modern vision transformers.
We explore the dynamics of dead neurons across a wide range of model widths and depths.
We test our approach on the benchmark CIFAR-10, CIFAR-100, SVHN, and Tiny Imagenet datasets, as well as their out-of-distribution corrupted variants.
When compared to standard ReLU networks, Synaptic Stripping significantly increased accuracy, reduced dead neurons and increased the number of active parameters on every experiment.
We see improvement with Synaptic Stripping over Gaussian Error Linear Units (GELU) on vision transformers, indicating that there is value in further investigations of our approach on large scale datasets like Imagenet or the JFT-300M \cite{Sun_2017_ICCV}.
Our experiments demonstrate the counter intuitive idea that removing parameters from a network can actually increase model capacity significantly, resulting in networks that are accurate, sparse, and efficient.

\small
\bibliographystyle{plain}
\bibliography{bibliography}

\end{document}